\documentclass[10pt, conference, a4paper]{IEEEtran}
\usepackage[table]{xcolor}
\usepackage{multirow}
\usepackage{orcidlink}
\usepackage{booktabs}
\usepackage{cite}
\usepackage{amsmath,amssymb,amsfonts}
\usepackage[nolist,nohyperlinks]{acronym}
\usepackage[inline]{enumitem}
\usepackage[detect-weight=true, detect-family=true]{siunitx}
\usepackage{algpseudocode}
\usepackage{algorithm}
\usepackage{svg}
\usepackage{multirow}
\usepackage{balance}
\usepackage{graphicx}
\usepackage[nameinlink, capitalise]{cleveref}
\usepackage[usestackEOL]{stackengine}
\DeclareSIUnit\ohm{\ensuremath\Omega}
\DeclareSIUnit\db{dB}
\DeclareSIUnit{\belmilliwatt}{Bm}
\DeclareSIUnit{\bel}{B}
\DeclareSIUnit{\bitpersecond}{bps}
\DeclareSIUnit{\samplepersecond}{Sps}
\DeclareSIUnit{\nothing}{\relax}
\usepackage[percent]{overpic}
\usepackage[export]{adjustbox}

\usepackage{pifont}
\DeclareRobustCommand{\IEEEauthorrefmark}[1]{\smash{\textsuperscript{\footnotesize #1}}}

\makeatletter
\newcommand*{\org@overidelabel}{}
\let\org@overridelabel\@verridelabel
\@ifpackagelater{acronym}{2015/03/21}{
  \renewcommand*{\@verridelabel}[1]{%
    \@bsphack
    \protected@write\@auxout{}{\string\AC@undonewlabel{#1@cref}}%
    \org@overridelabel{#1}%
    \@esphack
  }%
}{
  \renewcommand*{\@verridelabel}[1]{%
    \@bsphack
    \protected@write\@auxout{}{\string\undonewlabel{#1@cref}}%
    \org@overridelabel{#1}%
    \@esphack
  }%
}

\newcommand{\linebreakand}{%
  \end{@IEEEauthorhalign}
  \hfill\mbox{}\par
  \mbox{}\hfill\begin{@IEEEauthorhalign}
}
\makeatother

\begin{acronym}
    \acro{A-TDOA}{asynchronous \ac{TDOA}}
    \acro{AC}{Alternating Current}
    \acro{ADC}{Analog to Digital Converter}
    \acro{ADS-TWR}{alternative \acl{DS-TWR}}
    \acro{AI}{Artificial Intelligence}
    \acro{AP-TWR}{active-passive \ac{TWR}}
    \acro{AR}{Augmented Reality}
    \acro{ASK}{Amplitude Shift Keying}
    \acro{AUV}{Autonomous Underwater Vehicles}
    \acro{AUV}{Autonomous Underwater Vehicle}
    \acro{AoA}{Angle of Arrival}
    \acro{BAN}{Body Area Network}
    \acro{BC-EKF}{Baseline Constraint Extended Kalman Filter}
    \acro{BLE}{Bluetooth Low Energy}
    \acro{BLE}{Bluetooth Low Energy}
    \acro{BOM}{Bill of Material}
    \acro{CC-SS-TWR}{\acl{CFO}-corrected \ac{SS-TWR}}
    \acro{CFO}{carrier frequency offset}
    \acro{CIR}{channel-impulse response}
    \acro{CSMA}{Carrier Sense Multiple Access}
    \acro{DAQ}{Data Acquisition}
    \acro{DL-TDOA}{downlink \acl{TDOA}}
    \acro{DS-TWR}{double-sided two-way ranging}
    \acro{ECG}{Electrocardiogram}
    \acro{EKF}{Extended Kalman Filter}
    \acro{EQS-HBC}{Electro-Quasistatic Human Body Communication}
    \acro{EQS}{Electro-Quasistatic}
    \acro{FSK}{Frequency Shift Keying}
    \acro{FSR}{Force Sensing Resistor}
    \acro{FWR}{Full-Wave Rectifier}
    \acro{GBP}{Gaind-Bandwidth Product}
    \acro{GNSS}{Global Navigation Satellite System}
    \acro{GPIO}{General Purpose Input/Output}
    \acro{GUS}{global \ac{UWB} system}
    \acro{HBC}{Human Body Communication}
    \acro{HCI}{Human-Computer Interaction}
    \acro{HMI}{Human-Machine Interaction}
    \acro{IBPT}{Intra-Body Power Transfer}
    \acro{ICNIRP}{International Commission on Non-Ionizing Radiation Protection} 
    \acro{IC}{Integrated Circuit}
    \acro{IIR}{Infinite Impulse Response}
    \acro{IMU}{Intertial Measurement Unit}
    \acro{IoB}{Internet of Bodies}
    \acro{IoT}{Internet of Things}
    \acro{IoT}{Internet of Things}
    \acro{IoUT}{Internet of Underwater Things}
    \acro{KB}{Kilobyte}
    \acro{LoRaWAN}{Long Range Wide Area Network}
    \acro{LoRaWAN}{Long Range Wide Area Network}
    \acro{LoRa}{Long Range}
    \acro{LoS}{Line-of-Sight}
    \acro{LiPo}{Lithium-Polymer}
    \acro{MCU}{Microcontroller}
    \acro{ML}{Machine Learning}
    \acro{ML}{Machine Learning}
    \acro{NB}{Narrowband Communication}
    \acro{NFC}{near Field Communication}
    \acro{NVCR}{Negative Voltage Converter Rectifier}
    \acro{NVC}{Negative Voltage Converter}
    \acro{Nb-IoT}{Narrowband IoT}
    \acro{OOK}{On-Off Keying}
    \acro{PAN}{Personal Area Network}
    \acro{PCB}{Printed Circuit Board}
    \acro{PCE}{power Conversion Efficiency}
    \acro{PMIC}{Power Management IC}
    \acro{POM}{Polyoxymethylene}
    \acro{PULP}{Parallel Ultra-Low Power}
    \acro{PZT}{Lead Zirconium Titanate}
    \acro{QoS}{Quality of Service}
    \acro{RFID}{Radio Frequency Identification}
    \acro{RF}{Radio Frequency}
    \acro{RMSE}{root mean square error}
    \acro{RSSI}{Received Signal Strength Indicator}
    \acro{RSS}{Received Signal Strength}
    \acro{RTLS}{real-time locating system}
    \acro{RTOS}{Real-Time Operating System}
    \acro{RTT}{round-trip time}
    \acro{SF}{Spreading Factor}
    \acro{SS-TWR}{single-sided two-way ranging}
    \acro{SiP}{System in Package}
    \acro{SoC}{System on Chip}
    \acro{SoTA}{State-of-The-Art}
    \acro{SpO2}{Oxigen Saturation}
    \acro{SpO2}{Oxigen Saturation}
    \acro{TCDM}{Tightly Coupled Data Memory}
    \acro{TDMA}{time-division multiple-access}
    \acro{TDOA}{time difference of arrival}
    \acro{TIA}{Transimpedance Amplifier}
    \acro{TWR}{two-way ranging}
    \acro{ToF}{Time of Flight}
    \acro{UAC}{Underwater Acoustic Channel}
    \acro{UUID}{Universal Unique Identifier}
    \acro{UL-TDOA}{uplink \acl{TDOA}}
    \acro{UWB}{Ultra-Wideband}
    \acro{UWB}{Ultra-Wideband}
    \acro{UWN}{Underwater Wireless Network}
    \acro{UWSN}{Underwater Wireless Sensor Node}
    \acro{VR}{Virtual Reality}
    \acro{WBAN}{Wireless Body Area Network}
    \acro{WBAN}{Wirelss Body Area Network}
    \acro{WLAN}{Wireless Local Area Network}
    \acro{WPAN}{Wireless Personal Area Network}
    \acro{WPT}{Wireless Power Transfer}
    \acro{WP}{Work Package}
    \acro{nLoS}[NLOS]{non-Line-of-Sight}
    \acro{3D}{three-dimensional}
    \acro{2D}{two-dimensional}
    \acro{KPI}{key performance indicator}
\end{acronym}

          \usepackage{fancyhdr}   

\fancypagestyle{firstpage}{     
  \fancyhf{}
  \chead{\textcolor{gray}{This article has been accepted for publication in the conference of \\ the 10th IEEE International Workshop on Advances in Sensors and Interfaces.}}
  \fancyfoot[C]{\small{\textcolor{gray}{~\copyright~ 2025 IEEE.  Personal use of this material is permitted.  Permission from IEEE must be obtained for all other uses, in any current or future media, including reprinting/republishing this material for advertising or promotional purposes, creating new collective works, for resale or redistribution to servers or lists, or reuse of any copyrighted component of this work in other works.}}}
}

\begin{document}
\title{AniTrack: A Power-Efficient, Time-Slotted and Robust \acs{UWB} Localization System for Animal Tracking in a Controlled Setting\\}

\author{\IEEEauthorblockN{
Victor Luder\IEEEauthorrefmark{1}\orcidlink{0009-0005-4037-4220},
Lukas Schulthess\IEEEauthorrefmark{1}\orcidlink{0000-0002-6027-2927},
Silvano Cortesi\IEEEauthorrefmark{1}\orcidlink{0000-0002-2642-0797},
Leyla Rivero Davis\IEEEauthorrefmark{2}\orcidlink{0000-0002-0685-0976}, 
Michele Magno\IEEEauthorrefmark{1}\orcidlink{0000-0003-0368-8923}}\\
\IEEEauthorblockA{\IEEEauthorrefmark{1}\textit{Department of Information Technology and Electrical Engineering, ETH Zurich, Switzerland}}
\IEEEauthorblockA{\IEEEauthorrefmark{2}\textit{Zoo Zurich, Dept. for Research and Species Conservation, Switzerland}}
}
\maketitle
\begin{abstract}
Accurate localization is essential for a wide range of applications, including asset tracking, smart agriculture, and animal monitoring.
While traditional localization methods, such as \ac{GNSS}, Wi-Fi, and \ac{BLE}, offer varying levels of accuracy and coverage, they have drawbacks regarding power consumption, infrastructure requirements, and deployment flexibility.
\ac{UWB} is emerging as an alternative, offering centimeter-level accuracy and energy efficiency, especially suitable for medium to large field monitoring with capabilities to work indoors and outdoors.
However, existing \ac{UWB} localization systems require infrastructure with mains power to supply the anchors, which impedes their scalability and ease of deployment. This underscores the need for a fully battery-powered and energy-efficient localization system.

This paper presents an energy-optimized, battery-operated \ac{UWB} localization system that leverages \ac{LoRaWAN} for data transmission to a server backend. 
By employing \ac{SS-TWR} in a time-slotted localization approach, the power consumption both on the anchor and the tag is reduced, while maintaining high accuracy.
With a low average power consumption of $\mathbf{20.44\mskip3mu}$mW per anchor and $\mathbf{7.19\mskip3mu}$mW per tag, the system allows fully battery-powered operation for up to 25 days, achieving average accuracy of $\mathbf{13.96\mskip3mu}$cm with self-localizing anchors on a $\mathbf{600\mskip3mu m^2}$ testing ground.

To validate its effectiveness and ease of installation in a challenging application scenario, ten anchors and two tags were successfully deployed in a tropical zoological biome where they could be used to track Aldabra Giant Tortoises (\textit{Aldabrachelys gigantea}).
\end{abstract}

\acresetall

\section{Introduction}
\thispagestyle{firstpage} 
Monitoring animal behavior is essential for understanding their social dynamics, habitat preferences, and daily activity patterns~\cite{liptovszky_2024}. Not only do such findings advance scientific knowledge, but they also help to improve animal welfare~\cite{binding_2020} -- especially in controlled habitats like zoos~\cite{peter2025smartfeedingstationnoninvasive}.
Precise localization plays a key role in supporting these monitoring efforts~\cite{shannon_2024}, providing detailed data on their positions and interactions over time. 

Currently, a variety of localization technologies exist, such as satellite-based \ac{GNSS}, Wi-Fi, and \ac{BLE} localization~\cite{Obeidat_2021}. Although each system offers specific advantages, they also exhibit inherent limitations.
\ac{GNSS} offers wide-area coverage but has high energy requirements and struggles to provide reliable localization in dense urban areas or indoor applications due to signal degradation~\cite{smartcity_schulthess_2023}. 
On the other hand, Wi-Fi and \ac{BLE} are more suitable mainly for indoor \ac{RTLS}, where the installation of anchor infrastructure is feasible \cite{Dai_2023, Milano_2024}.
Moreover, their localization accuracy remains limited to a range of meters.
More precisely, Wi-Fi has been shown to achieve \qty{2.4}{\meter} accuracy \cite{Shen_2019}, while \ac{BLE} reaches accuracies around \qty{0.5}{\meter} in controlled environments\cite{cortesi25_proxim_based_approac_dynam_match}.

To address these limitations, the IEEE standard for \ac{UWB}, IEEE 802.15.4z \cite{IEEE_802.15.4z}, is emerging as a promising alternative.
\ac{UWB} offers sub \qty{20}{\centi\meter} accuracy \cite{Heinrich_2023}, high robustness against interference, and reliable performance in complex environments, making it significantly more accurate than Wi-Fi- or \ac{BLE}-based solutions.

To perform \ac{UWB} localization, various approaches can be implemented, such as localization through \ac{AoA} measurements~\cite{margiani23_angle_arriv_centim_distan_estim}, \ac{TDOA}~\cite{Mohanty_2023}, \ac{ToF}-based methods using either \ac{SS-TWR} or \ac{DS-TWR} ~\cite{Fakhoury_2023}, or any combination of those~\cite{cortesi25_wakel}.
However, \ac{TDOA} techniques require precise time synchronization among anchors, which adds complexity to the system. Similarly, \ac{AoA}-based localization demands sophisticated antenna arrays and calibration, while still relying on fixed anchor infrastructure. 
In contrast, \ac{SS-TWR} and \ac{DS-TWR} offer a more straightforward implementation without precise anchor synchronization. 
Independent of the chosen approach, \ac{UWB} localization systems rely on a network of anchors, typically involving a time-intensive manual deployment process.
While different solutions focused on lowering the power consumption of tags \cite{zhao21_uloc}, the anchors typically remain continuously active to handle localization requests from mobile tags.
This high power demand on the anchor's side requires a constant power supply and represents a fundamental limitation for system deployment.

Additionally, the system must be integrated into a communication network, allowing tags and anchors to upload position data to a server efficiently.
As a result, communication plays a key role in power consumption.
With the rise of \ac{IoT}, \ac{LoRaWAN} has been proven to bridge this gap, allowing power-efficient, reliable, and scalable data collection~\cite{smartcity_schulthess_2023}. 

This paper presents the design and implementation of a power-optimized and self-localizing \ac{UWB} \ac{RTLS} system, based on \ac{SS-TWR}.
The tag initiates the localization and follows a time-slotted approach, enabling positioning updates at intervals as short as \qty{10}{\second} with a \ac{2D} average accuracy of \qty{13.96}{\centi\meter}.
To ensure fast availability of localization data, the system integrates \ac{LoRaWAN} for energy-efficient data transmission. In particular, this article presents the following contributions:

\begin{itemize}
    \item Design and implementation of a hardware and software-based, power-optimized \ac{UWB} localization system with a \ac{2D} average accuracy of \qty{13.96}{\centi\meter} combined with real-time data upload through \ac{LoRaWAN}. 
    \item Introduction of a time-slotted localization schedule that reduces power consumption of our system to just \qty{20.44}{\milli\watt} for anchors and \qty{7.19}{\milli\watt} for tags at a localization interval of \qty{40}{\second}, enabling fully battery-powered operation for 25 days.
    \item A scalable system architecture, tested and characterized over an area of \qty{600}{\square\meter} using five anchors. 
    \item Successful field deployment, monitoring the real-time location of tags in a zoological setting.
\end{itemize}
\section{Related Work}\label{related_work}
The majority of \ac{UWB}-based localization techniques depend on its capability to measure time precisely.
An overview of such systems is provided in~\cref{tab:rel-works}.
The two primary and most widely adopted~\cite{coppens22_overv_uwb_stand_organ_ieee} methods for determining position exploiting time-of-flight information are \ac{TWR} and \ac{TDOA}.

\begin{table*}[htpb!]
    \begin{center}
    \caption{Comparison of state-of-the-art \ac{UWB} localization schemes.}\label{tab:rel-works}
    \vspace{-0.5cm}
    \resizebox{\textwidth}{!}{
    \renewcommand{\arraystretch}{1.3}
    \begin{tabular}{@{}llrrrrrrrrrrr@{}}
    \toprule
         &  \phantom{abc}&\textsc{ULoc}~\cite{zhao21_uloc}&\phantom{abc}&IROS'15~\cite{ledergerber15}&\phantom{abc}&\textsc{FlexTDOA}~\cite{patru23_flext}&\phantom{abc}&\textsc{GUS}~\cite{bach24_global_uwb_system}&\phantom{abc}&\textbf{This work}\\
      \midrule
      Anchor setup &&  Manual && Manual && Self-Loc && Manual && Self-Loc \\
      Loc. method &&   \acs{UL-TDOA} && \acs{DL-TDOA} && \acs{DL-TDOA} && \acs{DS-TWR} && \acs{SS-TWR} \\
      Tag access scheme &&  \acs{TDMA} && Simultaneous && Simultaneous && Random\(^a\) && \acs{TDMA} \\
      Anchor power supply &&   mains && -- && battery && -- && battery \\
             &&   -- && -- && \qty{400}{\milli\watt}@\qty{50}{\hertz} && -- && \qty{20.44}{\milli\watt}@\qty{0.025}{\hertz} \\
      Tag power supply &&   battery && -- && battery && -- && battery \\
             &&   \qty{31}{\micro\joule}\(^b\) && -- && \qty{400}{\milli\watt}@\qty{50}{\hertz} && -- && \qty{7.19}{\milli\watt}@\qty{0.025}{\hertz} \\
      Accuracy   && \qty{3.6}{\centi\meter} (3D) && \qty{14}{\centi\meter} (2D), \qty{28}{\centi\meter} (3D)&& \qty{17}{\centi\meter} (3D) && \qty{7.14}{\centi\meter} (2D) &&  \qty{13.96}{\centi\meter} (2D)\\
    \bottomrule
    \end{tabular}
    }
    \end{center}
    \hspace*{0.15cm}{\footnotesize \(^a\) no information on handling multiple tags.}\\
    \hspace*{0.15cm}{\footnotesize \(^b\) per localization. Sleep consumption is not presented.}
     \vspace{-0.5cm}
\end{table*}

In \ac{TWR}~\cite{dotlic18_rangin_method_utiliz_carrier_frequen_offset_estim}, the distance between two \ac{UWB} devices (typically a mobile tag and a fixed anchor) is estimated by exchanging time-stamped messages.
By measuring the \ac{RTT} of these messages, the system can derive the signal's \ac{ToF} and thus compute the distance. 
Since \ac{TWR} does not need device synchronization, it is computationally efficient and simple to implement. However, as the number of tags increases, energy consumption rises due to the increased amount of exchanged messages.

\ac{TDOA}~\cite{gust03positioning_tdoa}, on the other hand, estimates the position of the tag by looking at the variations in arrival times of transmitted messages from multiple synchronized anchors to a single tag. Because the anchors are time-synchronized and their positions are known, the differences in \ac{ToF} form a system of hyperbolic equations. \ac{TDOA} enables more scalable systems, as the tag transmits only once and is passive in the localization process. However, it requires sub-nanosecond synchronization accuracy between anchors and involves solving a non-linear least squares problem, making both implementation and computation more complex compared to \ac{TWR}.

In \cite{hindermann20_high_precis_real_time_locat}, Hindermann et al. developed and evaluated an \ac{UWB}-based localization system for cows and sows tracking in a real-world barn environment. Their proposed \ac{RTLS} is based on the \textsc{Qorvo DW1000} and uses an ALOHA random access protocol. Together with \ac{UL-TDOA}, it can support up to 100 tags at a \qty{1}{\second} localization interval with estimated packet collisions below 11.3\%. The overall localization error in static conditions was \qty{0.3}{\meter}.

With \textsc{ULoc}, Zhao et al. presented in~\cite{zhao21_uloc} an \ac{UWB}-based localization system capable of providing cm-accurate localization using \ac{3D}-\ac{AoA}. The system consists of eight \textsc{Qorvo DW1000} per anchor and a single one per tag. Similar to~\cite{hindermann20_high_precis_real_time_locat} \ac{UL-TDOA} is implemented, shifting the computational complexity to the infrastructure and thus reducing power consumption on the tag. \textsc{ULoc} achieves a static localization accuracy of \qty{3.6}{\centi\meter} and \qty{10}{\centi\meter} in a mobile tracking application.

In~\cite{ledergerber15}, Ledergerber et al. present a robot self-localization system similar to \ac{GNSS}.
Fixed anchors (with known location) broadcast \ac{UWB} signals, which are picked up by the robots to estimate their location.
Using \ac{TDOA}, the quadcopter accomplishes localization by exploiting the time difference of (almost) simultaneously transmitted signals of multiple anchors to the robot.
This passive approach makes the system intrinsically scalable, enabling several robots to localize simultaneously without central coordination. The system achieved a \ac{2D} \ac{RMSE} of \qty{14}{\centi\meter} and a \ac{3D} \ac{RMSE} of \qty{0.28}{\centi\meter} when tracking a quadcopters' trajectory. 

In~\cite{patru23_flext}, Patru et al. introduced \textsc{FlexTDOA}, a scalable \ac{RTLS} based on \ac{DL-TDOA}. In contrast to traditional \ac{TDOA}, \textsc{FlexTDOA} allocates a \ac{TDMA} schedule in which a distinct anchor initiates localization during each time slot, making the system scalable in the number of anchors.
An experimental evaluation based on five anchors shows a \ac{3D} localization accuracy of \qty{9}{\centi\meter} with a standard deviation of \qty{5}{\centi\meter}.
The passive nature of the tags enables scalability without augmenting system complexity.

Bach et al. \cite{bach24_global_uwb_system} present the \textsc{\ac{GUS}}, a localization system intended for high-accuracy positioning of autonomous mobile robots.
The system estimates heading angle and position by using two \ac{UWB} tags on the robot.
This method aims to prevent error accumulation by integrating information from \acp{IMU}.
To achieve this, they propose a tightly coupled \ac{EKF} fusion algorithm, which combines the fixed baseline constraint between the two tags with distance measurements using \ac{DS-TWR}.
The system improved the baseline \ac{RMSE} of \qty{8.6}{\centi\meter} using trilateration by 17.0\% down to \qty{7.14}{\centi\meter}.

In our work, \textsc{AniTrack} introduces a significant leap forward by combining the efficient \ac{SS-TWR}~\cite{dotlic18_rangin_method_utiliz_carrier_frequen_offset_estim} with self-localization concepts from~\cite{corbalan23_self_local_ultra_wideb_anchor}, where anchors autonomously determine their positions, and the \ac{TDMA}-based access scheme from \textsc{FlexTDOA}~\cite{patru23_flext}. 
This innovative combination not only simplifies the system architecture but also reduces power consumption, enabling a battery-powered deployment of up to 25 days for the anchors and tags with a localization interval of \qty{40}{\second}.
By integrating \ac{LoRaWAN} for data transmission, \textsc{AniTrack} ensures low-power operation while maintaining fast availability of real-time location data, thus pushing the boundaries of scalable, battery-operated, and efficient localization systems for animal tracking in controlled settings.
\section{AniTrack}\label{system_overview}
\begin{figure}[htpb!]
    \vspace{-0.1cm}
    \centering
    \includegraphics[width=0.9\linewidth]{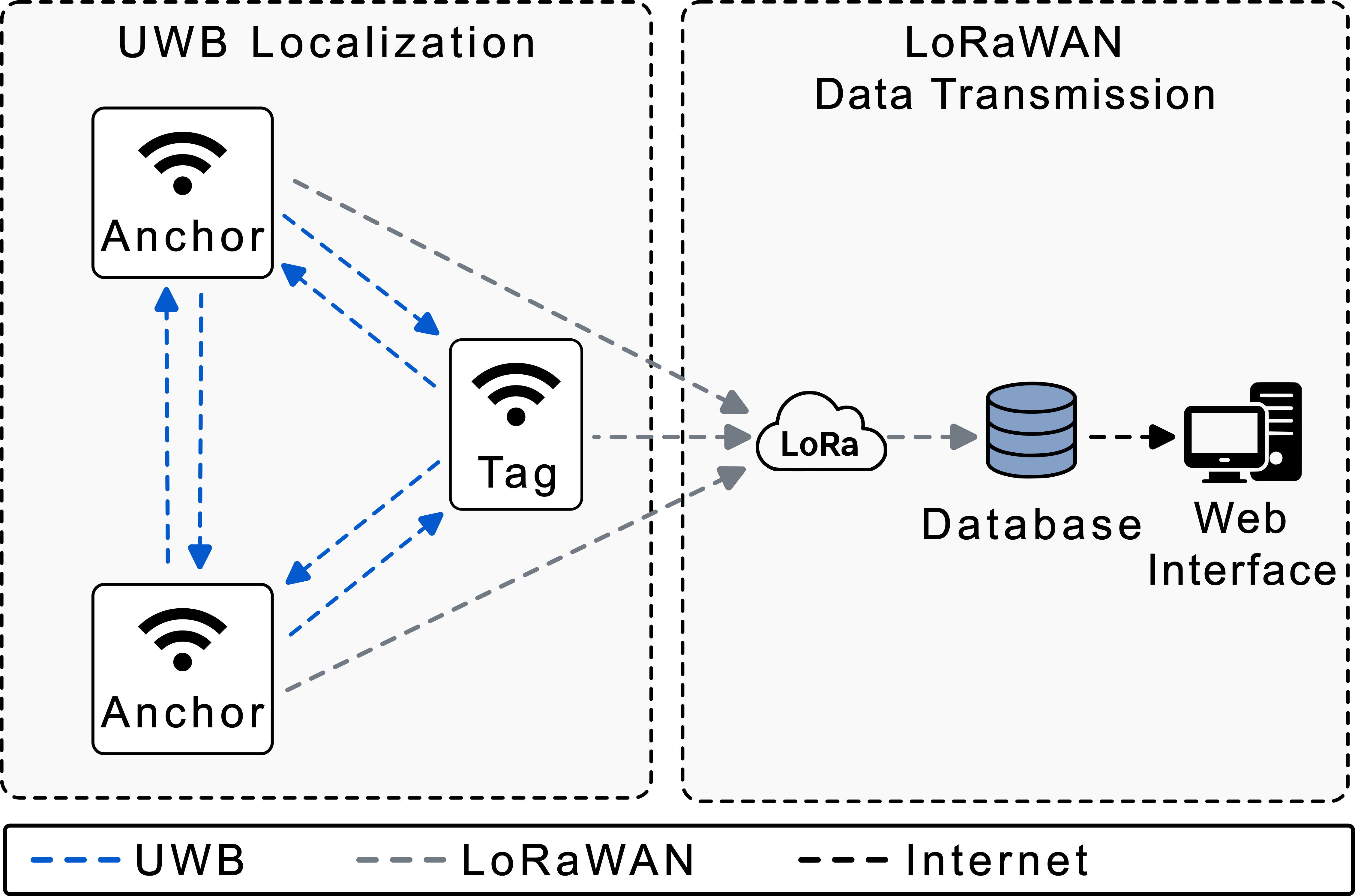}
    \vspace{-2mm}
    \caption{ \textsc{AniTrack} involves self-localizing anchors and tags which transmit their measured distances via \ac{LoRaWAN} to a database where coordinate calculations are performed.}
    \label{fig:data_path}
    \vspace{-0.45cm}
\end{figure}

The proposed system, as shown in~\cref{fig:data_path}, consists of custom hardware for anchor and tag, firmware for localization and sensor processing, which is complemented with a data collection setup and a database. 

In the first step, anchors and tags determine their distances relative to each other using \ac{UWB} and transmit this data over \ac{LoRaWAN} to a centralized database.
The \ac{LoRa} device is configured as a \textit{Class A} device, enabling uplinks at any time.
To maximize transmission range, the \ac{SF} is set to 12.
Once the data is stored in the database, it can be accessed over a web interface while the positions are calculated on the server side.
This enables real-time position monitoring, as sampled data is instantly available.

\subsection{Hardware}
\begin{figure}[htpb!]
    \vspace{-0.3cm}
    \centering
    \includegraphics[width=\linewidth]{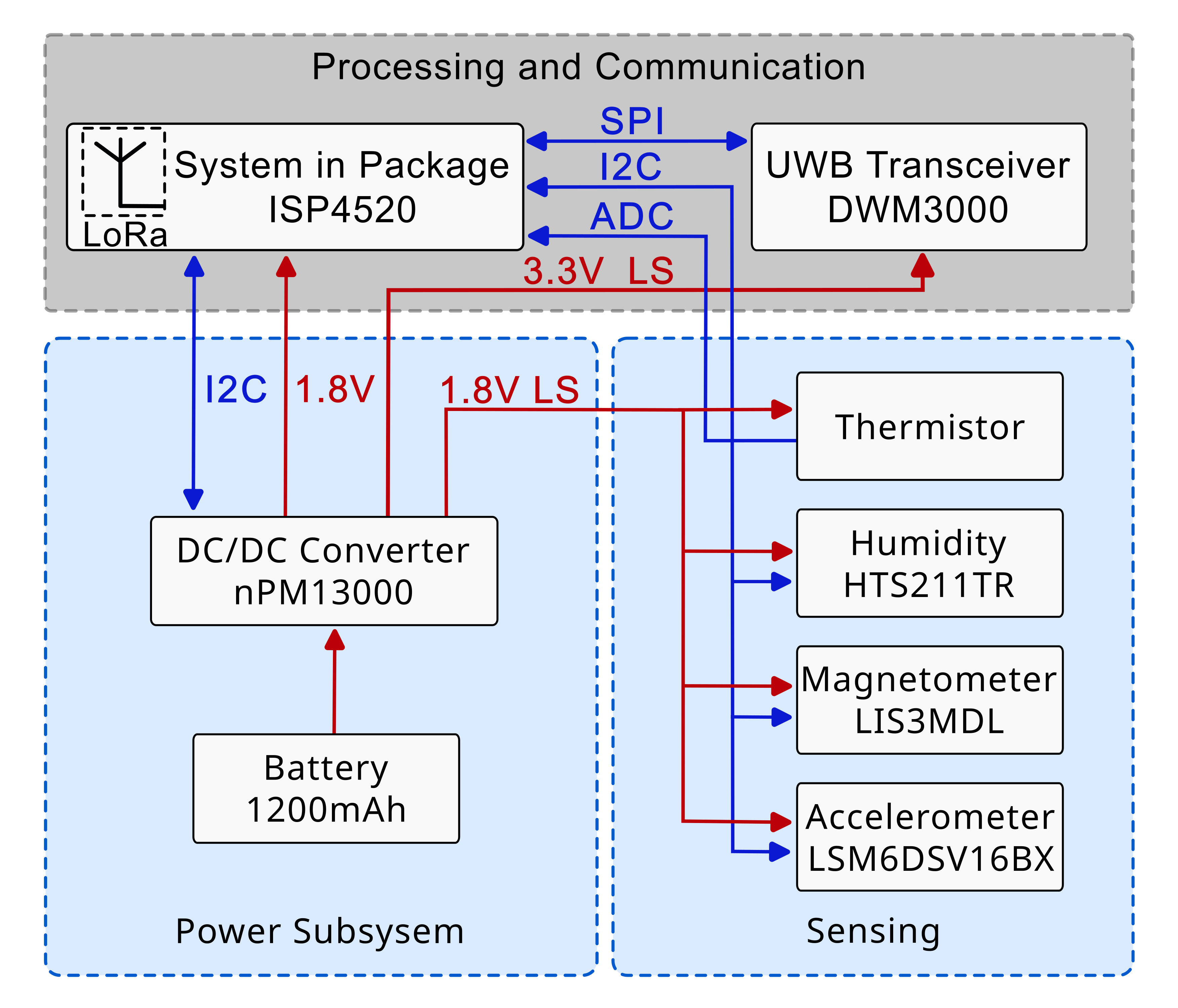}
    \vspace{-6mm}
    \caption{The hardware setup includes sensors, a \ac{UWB} transceiver, and a power management subsystem. The \ac{PMIC} supplies \qty{1.8}{\volt} and \qty{3.3}{\volt}, both voltages controlled by a load switch (LS).}
    \label{fig:hardware_setup}
    \vspace{-0.4cm}
\end{figure}

The anchors and tags share the same hardware and are built around the \textsc{ISP4520} \ac{SiP}, hosting an \textsc{nRF52832} \ac{MCU}, a \textsc{SX126x} \ac{LoRa} transmitter, and an integrated antenna, see \cref{fig:hardware_setup}.
An external \textsc{DWM3000} \ac{UWB} transceiver is used to handle the \ac{UWB} communication.
To optimize power consumption, the \textsc{nPM1300} \ac{PMIC} manages the battery, generates the system voltages, and controls power domains using the integrated load switches.
In our setup, all sensors operate at \qty{1.8}{\volt}, allowing them to be switched off when not in use. The \textsc{DWM3000} is powered by a separate \qty{3.3}{\volt} domain that can also be turned off.
The device is equipped with various sensors, including an IP67-certified thermistor for measuring temperature in harsh environments, an \textsc{HTS211TR} humidity sensor, a \textsc{LIS3MDL} magnetometer for orientation measurement, and the \textsc{LSM6DSV16BX} \ac{IMU} for movement recognition. 

\subsection{Localization Principle}\label{localization}
To do a \ac{2D} localization, the distance between the tag and at least 3 anchors is measured using the \ac{RTT} of a \ac{SS-TWR} exchange as described in~\cite{dotlic18_rangin_method_utiliz_carrier_frequen_offset_estim}.
The protocol is initiated by the tag, which utilizes the anchors' responses to compute the carrier-frequency-offset-corrected time-of-flight and finally estimate its distance to the anchors.
Due to inevitable measurement inaccuracies, gathering data from as many anchors as possible is crucial. The final position of the tag is then evaluated by solving the least squares estimate of \cref{eq:least_squares} for \( N \) anchors.

\begin{equation}
\mathbf{p}^* = \arg\min_{\mathbf{p}} \sum_{i=1}^{N} \left( \|\mathbf{p} - \mathbf{a}_i\| - d_i \right)^2
\label{eq:least_squares}
\end{equation}

where:
\begin{itemize}
    \item \( \mathbf{p} = (x, y) \) is the estimated position of the tag,
    \item \( \mathbf{a}_i = (x_i, y_i) \) is the known position of the \(i\)-th anchor,
    \item \( \mathbf{d_i} \) is the measured distance from the tag to the \(i\)-th anchor,
    \item \( \|\mathbf{p} - \mathbf{a}_i\| \) Euclidean distance between tag and anchor \(i\),
     \item \( \mathbf{N} \) is the number of anchors.
\end{itemize}

\subsection{Time-Slotted Localization}
\begin{figure}[htpb!]
 \vspace{-0.4cm}
    \centering
    \begin{overpic}[width=0.85\linewidth]{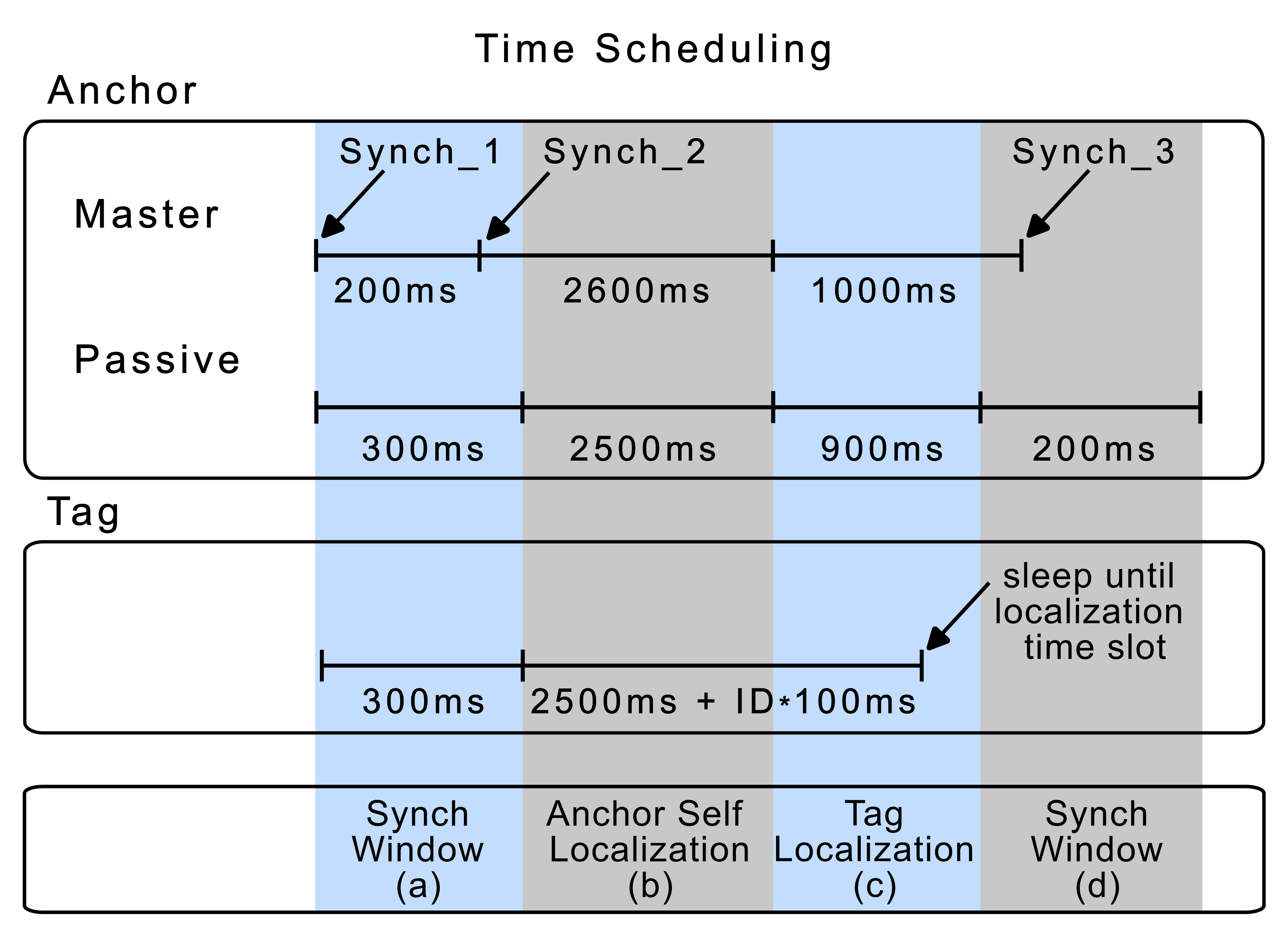}
        \put(-5,61){(1)}
        \put(-5,28){(2)}
    \end{overpic}
    \vspace{-0.3cm}
    \caption{The localization phase, for anchors (1) and tags (2), is divided into synchronization (a \& d), anchor self-localization (b), and tag localization (c). Localization happens sequentially depending on the given ID.}
    \label{fig:time_slots}
     \vspace{-0.2cm}
\end{figure}

\ac{SS-TWR} localization scheme relies on responsive anchors, ready to respond whenever a tag initiates a measurement request. Consequently, traditional approaches always have their anchors on, making this approach power-intensive.
Since our anchors are battery-powered, continuous operation significantly reduces runtime and increases maintenance efforts due to frequent battery replacements.  
To solve this issue, a time-slotted localization protocol is introduced.
It schedules specific time intervals when anchors are awake and ready to respond to tags, reducing overall power consumption.
The duration of these time slots depends on the number of anchors and tags in the system.  
For a setup with ten anchors and up to ten tags, a total active time of \qty{3900}{\milli\second} is proposed. This period is divided into two phases: Each anchor is allocated \qty{200}{\milli\second} for self-localization, while a tag has \qty{100}{\milli\second} for its localization. These time slots are illustrated in \cref{fig:time_slots} with dedicated slots for anchor self-localization and tag localization.

To achieve a reliable anchor synchronization, the anchors are categorized into master, relay, and passive anchors.
The master anchor initiates the time slots by broadcasting a synchronization message.
All other anchors use this to start and synchronize their timers.
To enhance reliability, a second synchronization message is sent by the master \qty{200}{\milli\second} after the first.
This allows already synchronized anchors to verify correct alignment, even if they missed the first message due to wake-up timing differences.
Finally, a third synchronization message is broadcast \qty{100}{\milli\second} before the end of the active phase. These messages are uniquely labeled, enabling passive anchors to verify the order of reception and ensure they remain in sync.  
Additionally, relay anchors extend the synchronization range by retransmitting these sync messages, ensuring scalability even outside of the master anchor’s communication range.

\begin{figure} [t!]
    \centering\includegraphics[width=0.9\linewidth]{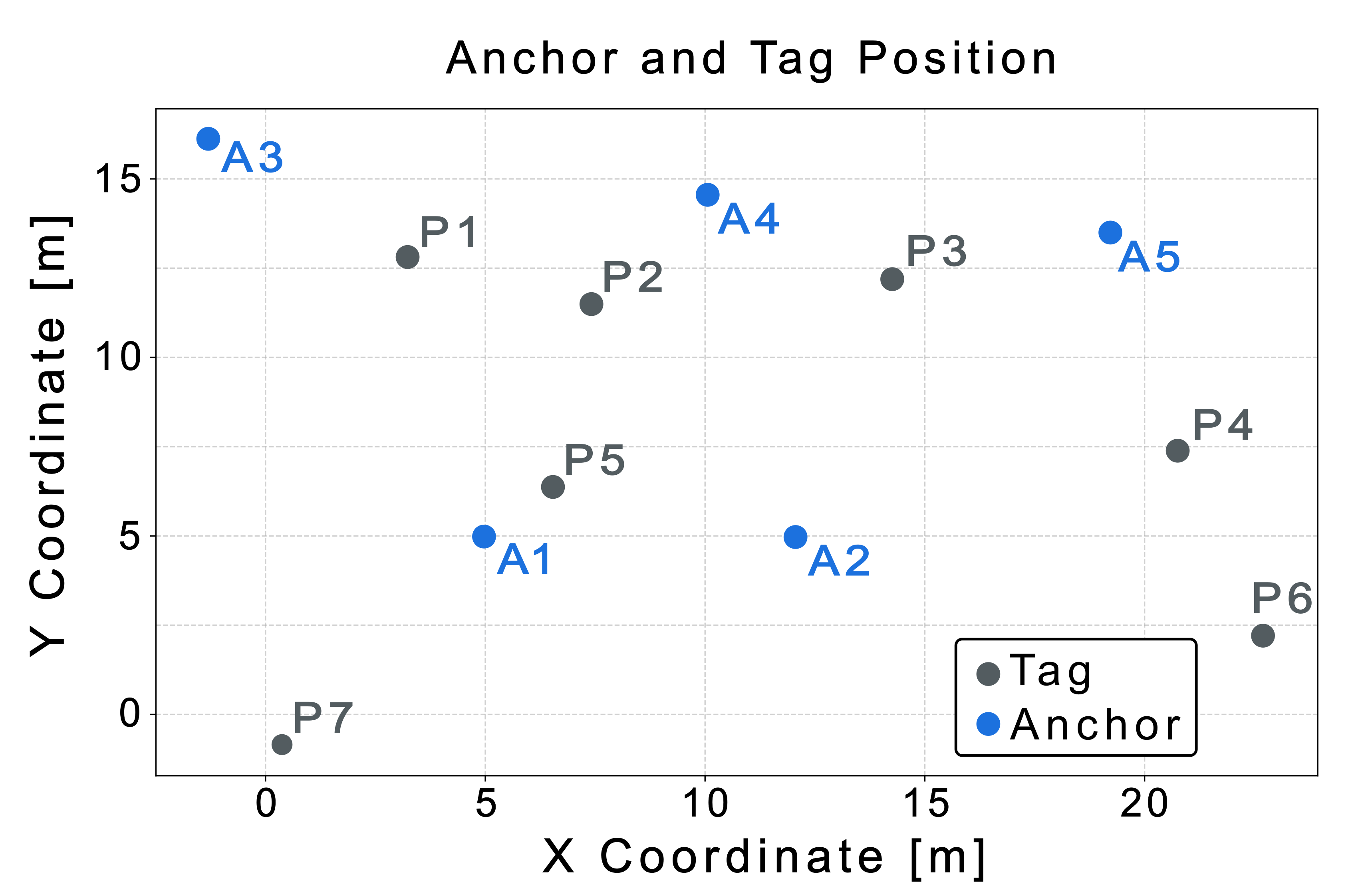}
    \vspace{-0.3cm}
    \caption{Accuracy evaluation in a test setup with five anchors and tag positions distributed across a \qty{600}{\square\meter} area.}
    \label{fig:anchor_tag_position}
    \vspace{-0.4cm}
\end{figure}

\subsection{Anchor Self-Localization}
The self-localization principle leverages the anchors' ability to measure distances between each other. This approach enables the system to continuously collect distance data between anchors. By designating one anchor as the origin together with an axis to a second anchor, a local \ac{2D} coordinate system can be established, determining all anchors' positions as described in \cref{localization}.
A key advantage of this method is its reliance on a large number of measurements. Since self-localization is performed during every active phase, each anchor regularly updates its position. 
This continuous monitoring allows the system to detect changes in anchor placement and automatically update their positions.
\section{System Characterization}\label{characterization}
The proposed system has been evaluated based on the \acp{KPI} localization accuracy and power consumption.

\subsection{Localization Accuracy}
The localization accuracy of the system is examined with a single tag (at seven different positions) and five anchors deployed on an open field (\qty{600}{\square\meter}) according to~\cref{fig:anchor_tag_position}. The ground truth positions were determined using a laser distance measurement device (\textsc{Bosch GLM150-27C}). 

To estimate the tags' location, we used the anchors' ground truth position and the anchors' self-localized position.
With the ground truth position, an average accuracy of \qty{16.57}{\centi\meter} and a standard deviation of \qty{7.86}{\centi\meter} was achieved. As shown in~\cref{localization_acuracies}, the estimations based on self-localized anchors were more accurate, achieving an accuracy of \qty{13.96}{\centi\meter} and a standard deviation of \qty{6.77}{\centi\meter}.

\begin{table}[htpb!]
    \vspace{-0.2cm}
  \renewcommand{\arraystretch}{1.3}
  \centering
  \caption{Statistics of the localization experiments, with ground truth anchor positions (GT) and self-localized anchor positions (SL). Values are in \unit{\centi\meter}.}\label{localization_acuracies}
  \begin{tabular}{@{}lrrrcrrr@{}}
    \toprule
    & \multicolumn{3}{c}{\textsc{GT Anchors}} & \phantom{a} & \multicolumn{3}{c}{\textsc{SL Anchors}} \\
    \cmidrule{2-4} \cmidrule{6-8}
    & \(x\) & \(y\) & 2D && \(x\) & \(y\) & 2D \\
    \midrule
    avg & 9.19 & 13.25 & 16.57 && 8.4 & 10.68 & 13.96 \\
    md & 7.71 & 12.8 & 14.62 && 7.9 & 7.92 & 10.98 \\
    \(\sigma(\cdot)\) & 5.82 & 6.73 & 7.86 && 5.12 & 5.6 & 6.77\\
    \bottomrule
  \end{tabular}
      \vspace{-0.3cm}
\end{table}

\subsection{Power Evaluation}
To assess the power consumption of the battery-powered tag and anchors, the individual system states -- including low-power idle mode, anchor self-localization, tag localization, and \ac{LoRa} transmission -- were measured using a \textsc{Power Profiler Kit II} from \textsc{Nordic Semiconductors}.

\begin{table}[b]
\renewcommand{\arraystretch}{1.3}
\centering
\caption{Power consumption of the whole system running at \qty{3.7}{\volt} for different system states and average consumption for a localization period of \qty{10}{\second}}\label{tbl:power_measurement}
\begin{tabular}{@{}llrr@{}} 
    \toprule
    &  & Power Consumption & Duration \\ 
    \midrule
    \multirow[t]{5}{*}{Anchor} & Sleep Mode & \qty{43.66}{\micro\watt} & -- \\
                            & Anchor Transceiver On & \qty{140.93}{\milli\watt} & \qty{3.74}{\second} \\
                            & Self-Localizing of Anchors & \qty{40.55}{\milli\watt} & \qty{156}{\milli\second} \\
                            & LoRa Transmission & \qty{67.82}{\milli\watt} & \qty{4.17}{\second} \\
                             & \textbf{Average (\(\mathbf{\qty{10}{\second}}\) period)} &\textbf{\(\mathbf{\qty{81.6}{\milli\watt}}\)} & \textbf{\(\mathbf{\qty{10}{\second}}\)} \\
    \midrule
    \multirow[t]{3}{*}{Tag} & Sleep Mode & \qty{43.66}{\micro\watt} & -- \\
                        & Tag Localization & \qty{52.32}{\milli\watt} & \qty{63}{\milli\second} \\
                         & LoRa Transmission & \qty{67.82}{\milli\watt} & \qty{4.17}{\second} \\
                         & \textbf{Average (\(\mathbf{\qty{10}{\second}}\) period)} & \textbf{\(\mathbf{\qty{28.6}{\milli\watt}}\)} &  \textbf{\(\mathbf{\qty{10}{\second}}\)} \\
    \bottomrule
\end{tabular}
    \vspace{-0.1cm}
\end{table}

\begin{figure}[htpb!]
    \vspace{-0.1cm}
    \centering \includegraphics[width=0.9\linewidth]{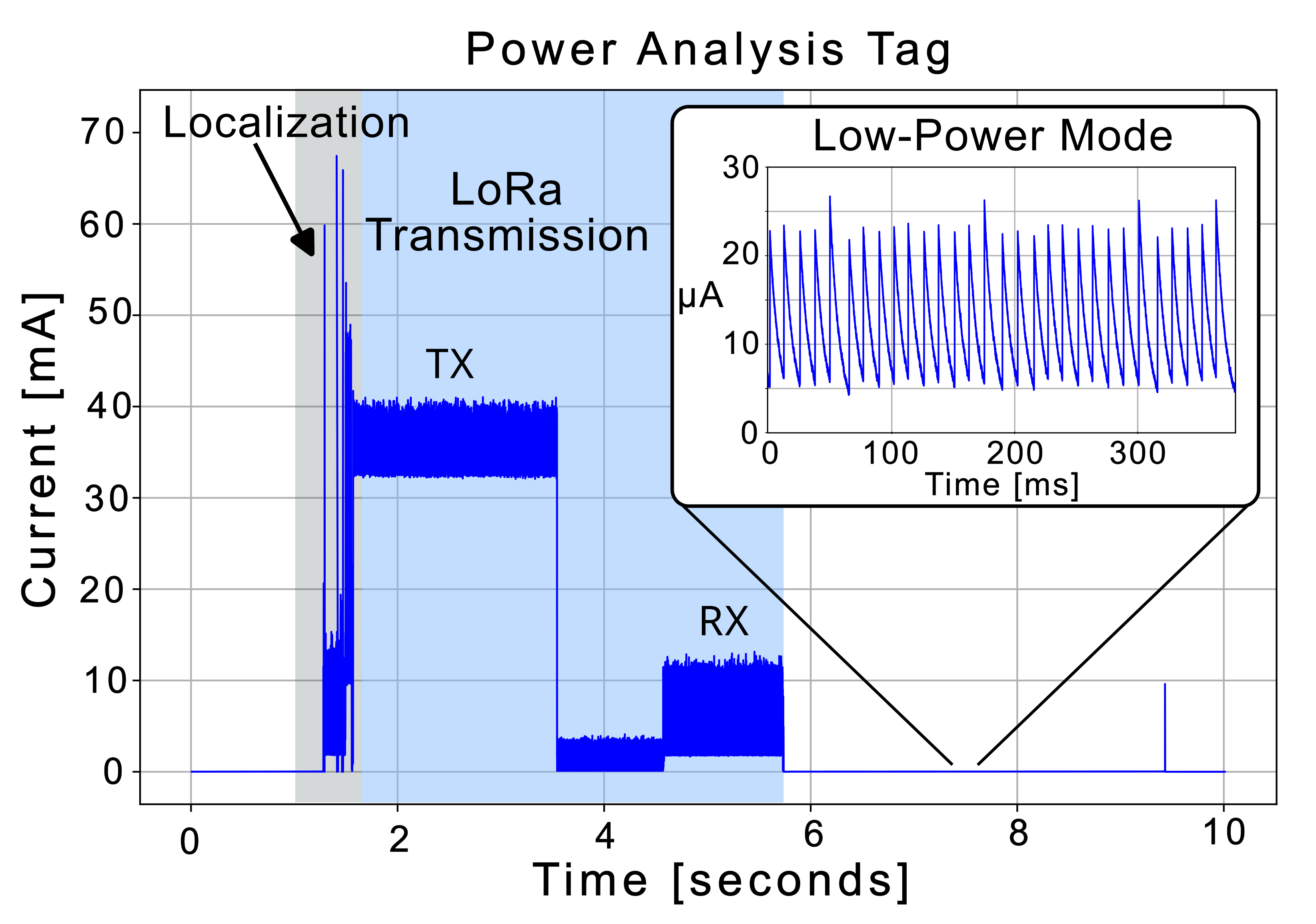}
        \vspace{-0.3cm}
    \caption{Power consumption of the tag during each phase: localization (light gray), \ac{LoRa} transmission (blue), and sleep mode.}
    \label{fig:power_tag}
        \vspace{-0.1cm}
\end{figure}

\begin{figure}[htpb!]
    \vspace{-0.2cm}
    \centering
    \includegraphics[width=0.9\linewidth]{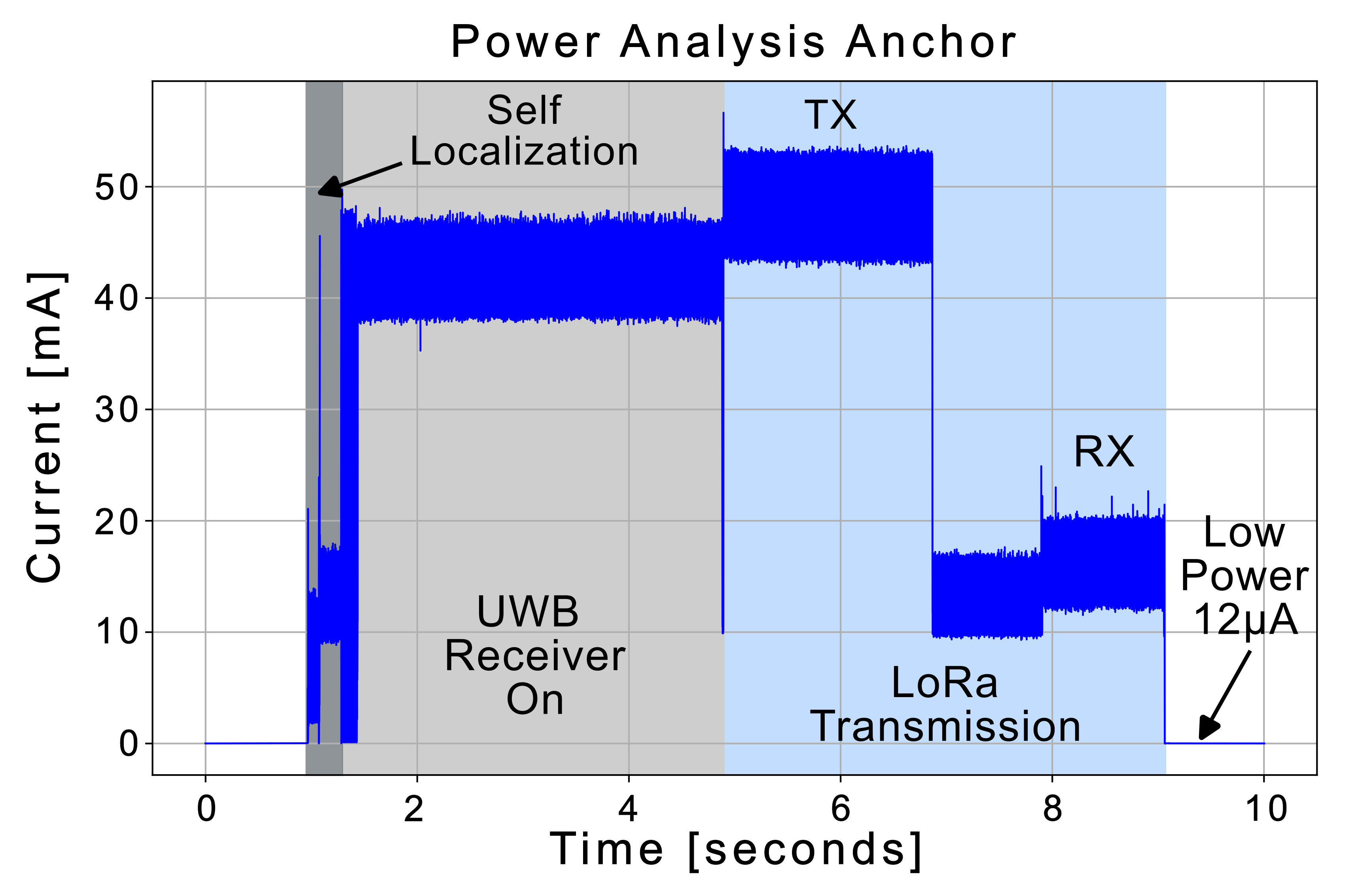}
    \vspace{-0.3cm}
    \caption{Power consumption of the anchor during each phase: self-localization (gray), tag-localization (light gray), LoRa transmission (blue), and sleep mode.}
    \label{fig:power_anchor}
    \vspace{-0.35cm}
\end{figure}

These measurements showed an average power consumption of \qty{81.6}{\milli\watt} for the anchors and \qty{28.6}{\milli\watt} for the tag. This assumes a high localization interval of \qty{10}{\second}.
The individual states and their current draw are shown in \cref{fig:power_anchor} and \cref{fig:power_tag}.
With a lower localization period of \qty{40}{\second}, the average power consumption can be lowered to \qty{20.44}{\milli\watt} for the anchors and \qty{7.19}{\milli\watt} for the tag.
If the anchors stop to upload data through \ac{LoRa} once they have localized themselves, their power consumption is as low as \qty{13.38}{\milli\watt}. This extends the battery runtime to 29.96 days for the anchors and 25.7 days for the tags, given a \ac{LiPo} battery size of \qty{2600}{\milli\ampere{}\hour} for the anchors and \qty{1200}{\milli\ampere{}\hour} for the tag. 

\subsection{System Evaluation}\label{field_application}
To evaluate the system's performance, including backend data collection, the setup concept described was tested in a real-world application by deploying a temporary sensor that could be used for Aldabra Giant Tortoises in the semi-natural Masoala rainforest enclosure at Zoo Zurich. 

\begin{figure}[htpb!]
        \vspace{-0.1cm}
    \centering
    \includegraphics[width=0.9\linewidth]{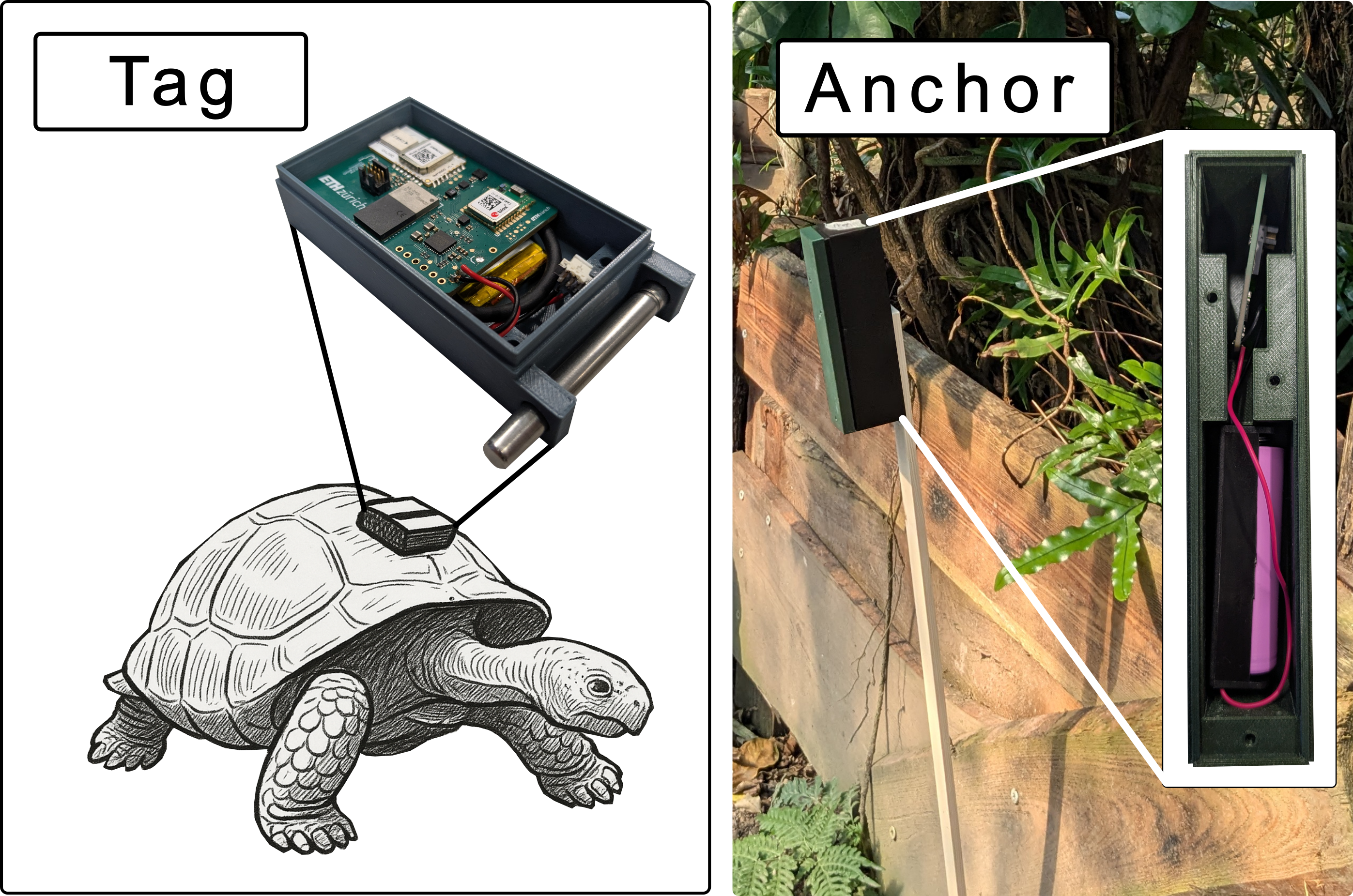}
        \vspace{-0.3cm}
    \caption{Possible real-world application; monitoring the location of Aldabra Giant Tortoises.}
    \label{fig:final_setup}
        \vspace{-0.3cm}
\end{figure}

\begin{figure}[htpb!]
    \centering
    \includegraphics[width=0.9\linewidth]{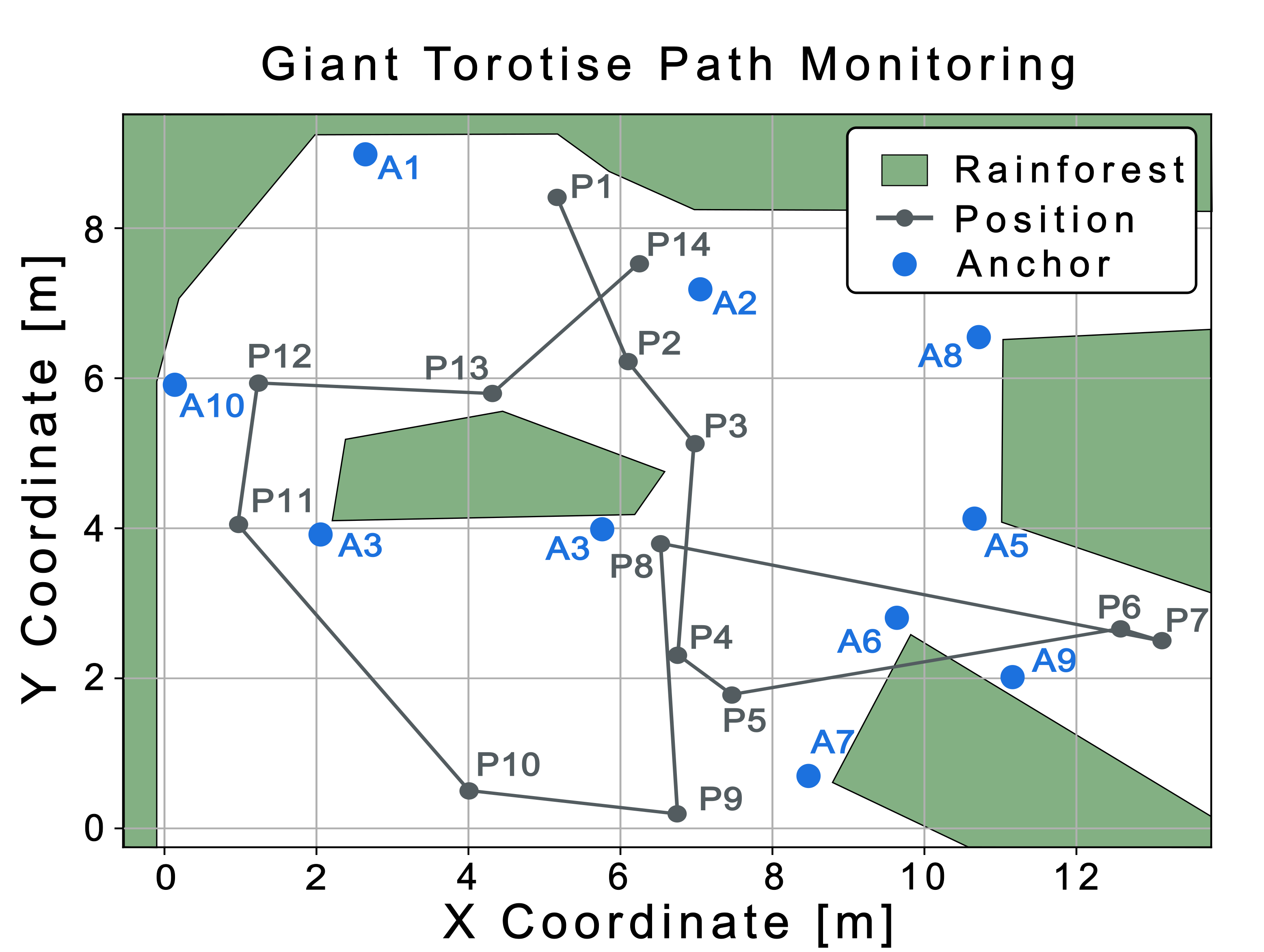}
        \vspace{-0.3cm}
    \caption{Tracking position within the enclosure by simulating the behavior of an Aldabra Giant Tortoise during the two-hour trial period.}
    \label{fig:tortoise_path_monitoring}
        \vspace{-0.4cm}
\end{figure}

The described system was installed with a position collecting interval of \qty{40}{\second}, acquiring data for a two-hour period.  
To ensure a flexible localization setup, the anchors are mounted on \qty{1.5}{\meter} wooden poles for easy positioning, as shown in \cref{fig:final_setup}. Additionally, two tags are moved within the tropical biome, simulating the tortoises moving.
The final setup of how it could be installed on the animals is illustrated in \cref{fig:final_setup}, while \cref{fig:tortoise_path_monitoring} presents a selection of the collected position data. The monitored enclosure covers a total area of approximately \qty{126}{\square\meter}.

\section{Conclusion}\label{conclusion}
While conventional localization systems setups may provide precise localizations, they often come with specific requirements such as time-consuming anchor installation, manual calibration, and a mains power connection.
To address this challenge, this paper presents a power-efficient and fully battery-powered \ac{UWB} localization system that enables precise localization with minimal setup overhead.
A time-slotted \ac{UWB} \ac{TWR} localization approach reduces the power consumption while preserving the flexibility of a battery-powered setup.
The suggested approach achieves an average \ac{2D} localization accuracy of \qty{13.96}{\centi\meter} on \qty{600}{\square\meter}.
When setting the localization interval to \qty{40}{\second}, the system achieves an average power consumption of \qty{7.19}{\milli\watt} for tags and \qty{20.44}{\milli\watt} for anchors, enabling a battery runtime of 25 days for the entire system.
Furthermore, the full end-to-end implementation allows for transmitting collected data over \ac{LoRaWAN} to the server backend and a directly accessible web interface.
A semi-permanent deployment in a zoological setting proved the system's functionality and robustness.
Future work will focus on extended long-term testing, expanding data collection, and further exploration of the system’s capabilities.
\section{Acknowledgments}
We extend thanks for support of Zoo Zurich's Masoala Rainforest team:  Basil von Ah, Angus Sünner, Francesco Biondi, and Daniela Kollmuss, 
as well as EWZ (Elektrizitätswerk der Stadt Zürich) for providing \ac{LoRaWAN} infrastructure and to Marcus Cathomen for his generous support and collaboration initiation.

\bibliographystyle{IEEEtran}
\bibliography{bib/references.bib}

\end{document}